\documentclass[10pt,twocolumn,letterpaper]{article}

\usepackage[pagenumbers]{cvpr} 

\makeatletter
\@namedef{ver@everyshi.sty}{}
\makeatother
\usepackage{tikz}

\usepackage{listings}
\usepackage{xcolor}
\definecolor{eclipseStrings}{RGB}{42,0.0,255}
\definecolor{eclipseKeywords}{RGB}{127,0,85}
\colorlet{numb}{magenta!60!black}

\lstdefinelanguage{json}{
    basicstyle=\scriptsize\ttfamily,
    commentstyle=\color{eclipseStrings}, 
    stringstyle=\color{eclipseKeywords}, 
    numberstyle=\scriptsize,
    stepnumber=1,
    numbersep=8pt,
    showstringspaces=false,
    breaklines=true,
    string=[s]{"}{"},
    comment=[l]{:\ "},
    morecomment=[l]{:"},
    literate=
        *{0}{{{\color{numb}0}}}{1}
         {1}{{{\color{numb}1}}}{1}
         {2}{{{\color{numb}2}}}{1}
         {3}{{{\color{numb}3}}}{1}
         {4}{{{\color{numb}4}}}{1}
         {5}{{{\color{numb}5}}}{1}
         {6}{{{\color{numb}6}}}{1}
         {7}{{{\color{numb}7}}}{1}
         {8}{{{\color{numb}8}}}{1}
         {9}{{{\color{numb}9}}}{1}
}
\lstdefinelanguage{python}{
    basicstyle=\scriptsize\ttfamily,
    commentstyle=\color{eclipseStrings}, 
    stringstyle=\color{eclipseKeywords}, 
    numberstyle=\scriptsize,
    stepnumber=1,
    numbersep=8pt,
    showstringspaces=false,
    breaklines=true,
    string=[s]{"}{"},
    comment=[l]{:\ "},
    morecomment=[l]{:"},
    literate=
        *{0}{{{\color{numb}0}}}{1}
         {1}{{{\color{numb}1}}}{1}
         {2}{{{\color{numb}2}}}{1}
         {3}{{{\color{numb}3}}}{1}
         {4}{{{\color{numb}4}}}{1}
         {5}{{{\color{numb}5}}}{1}
         {6}{{{\color{numb}6}}}{1}
         {7}{{{\color{numb}7}}}{1}
         {8}{{{\color{numb}8}}}{1}
         {9}{{{\color{numb}9}}}{1}
}

\usepackage{graphicx}
\usepackage{amsmath}
\usepackage{amssymb}
\usepackage{booktabs}

\usepackage[pagebackref,breaklinks,colorlinks]{hyperref}

\usepackage[capitalize]{cleveref}
\Crefname{section}{Section}{Sections}
\Crefname{table}{Table}{Tables}
\crefname{table}{Tab.}{Tabs.}

\usepackage[numbers,sort]{natbib} 
\usepackage{tcolorbox} 

\def\cvprPaperID{*****} 
\def\confName{CVPR}
\def\confYear{2023}

\begin{document}

\title{The Casual Conversations v2 Dataset\\ {\small A diverse, large benchmark for measuring fairness and robustness in audio/vision/speech models}}

\author{
Bilal Porgali\\
{\tt\small porgali@meta.com}
\and
Vítor Albiero\\
{\tt\small valbiero@meta.com}
\and
Jordan Ryda\\
{\tt\small jryda@meta.com}
\and 
Cristian Canton Ferrer\\
{\tt\small ccanton@meta.com}
\and
Caner Hazirbas\\
{\tt\small hazirbas@meta.com}\\ \\
Meta AI
}

\maketitle

\begin{abstract}
   This paper introduces a new large consent-driven dataset aimed at assisting in the evaluation of algorithmic bias and robustness of computer vision and audio speech models in regards to 11 attributes that are self-provided or labeled by trained annotators.
   The dataset includes 26,467 videos of 5,567 unique paid participants, with an average of almost 5 videos per person, recorded in Brazil, India, Indonesia, Mexico, Vietnam, Philippines, and the USA, representing diverse demographic characteristics. 
   The participants agreed for their data to be used in assessing fairness of AI models and provided self-reported age, gender, language/dialect, disability status, physical adornments, physical attributes and geo-location information, while trained annotators labeled apparent skin tone using the Fitzpatrick Skin Type and Monk Skin Tone scales, and voice timbre.
   Annotators also labeled for different recording setups and per-second activity annotations. 
\end{abstract}

\section{Introduction}
Ethical considerations on dataset construction~\cite{hazirbas2022ccv2lit, jerone2023ethical} has become more significant in AI shortly after several number of studies carried out by researchers to identify fairness concerns of biometrics and facial processing technologies were published~\cite{buolamwini2018gender, grother2018ongoing, cook2019demographic, howard2019effect, drozdowski2020demographic, krishnapriya2020issues, albiero2020analysis}, and discussions of data protection and related regulation have continued to evolve~\cite{whitehouse2022aibill, eu2023artificial}.

One of the biggest challenges of identifying fairness issues has been a lack of clean, and most importantly responsibly-constructed, benchmarks. 
Casual Conversations~\cite{hazirbas2022ccv1}, Dollar Street~\cite{rojas2022dollarstreet}, Open Images MIAP~\cite{schumann2021step}, FairFace~\cite{karkkainen2021fairface}, UTK Faces~\cite{zhifei2017age}, RFW~\cite{Wang_2019_ICCV}, and MORPH~\cite{ricanek2006morph},
are some of the most widely used datasets to identify fairness gaps~\cite{vries2019does, goyal2022fairness}. 
However, each of these has their own limitations. Casual Conversations (CCv1) is not geographically diverse (U.S. only), Dollar Street does not have person attributes such as age \& gender, Open Images MIAP uses perceived gender instead of self-provided, FairFace, UTK Faces, RFW are made up of images collected from the internet. MORPH has binary gender with limited number of attributes that are age, gender, race, height, weight, and eye coordinates.
Moreover, except Casual Conversations, all other aforementioned datasets are designed to measure only computer vision models.

\begin{figure}
    \centering
    \includegraphics[width=1\linewidth,clip,trim=13cm 3cm 13cm 2.6cm]{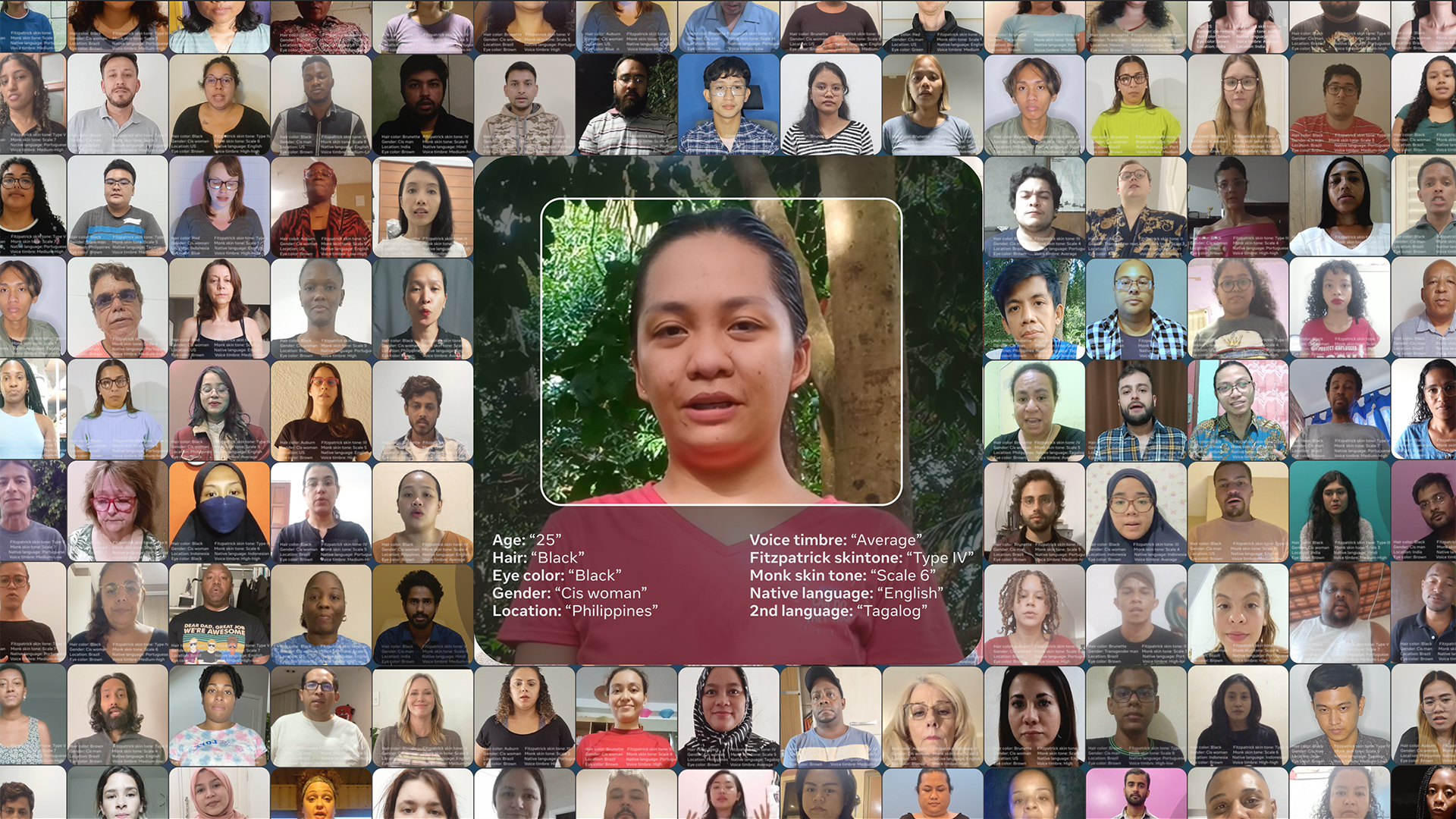}
    \caption{\textit{\textbf{The Casual Conversations v2}} dataset includes a total of 11 attributes that are self-provided or annotated.} 
    \label{fig:teaser}
    \vspace{-0.4cm}
\end{figure}

To make progress against aforementioned limitations, we propose a large, diverse and consent-driven audio/vision/speech dataset with many attributes,~\ie \textit{Casual Conversations v2}\footnote{\url{https://ai.facebook.com/datasets/casual-conversations-v2-dataset}}. Our dataset is composed of 26,467 videos of 5,567 unique paid participants recorded in seven countries and include 11 self-provided and annotated attributes.
Participants directly provided us their data to be used in AI and provided self-identified \textit{age, gender, language/dialect, disability, physical adornments \& attributes and geo-location} information. In addition, trained annotators labeled participants' \textit{apparent skin tones} \& \textit{voice timbre} and annotated videos for \textit{different recording setup} and per-second \textit{activity} on videos. Subcategories of these attributes were selected and defined based on the literature survey that is carried out by Hazirbas~\etal~\cite{hazirbas2022ccv2lit}.

To our knowledge, Casual Conversations v2 (CCv2) is the most comprehensive dataset that is constructed with ethical considerations in mind and may be used in many AI tasks from computer vision and audio/speech recognition to deepfake detection — not only for measuring fairness but also evaluating robustness of the models. Furthermore, this dataset may also be used for model training (note that certain labels age, gender, disability, physical adornments \& attributes may not be used for training). Please see the license agreement for further use of the data\footnote{\url{https://ai.facebook.com/datasets/casual-conversations-v2-downloads}}.

\section{Related Work}
\begin{figure}[t!]
    \includegraphics[width=\columnwidth]{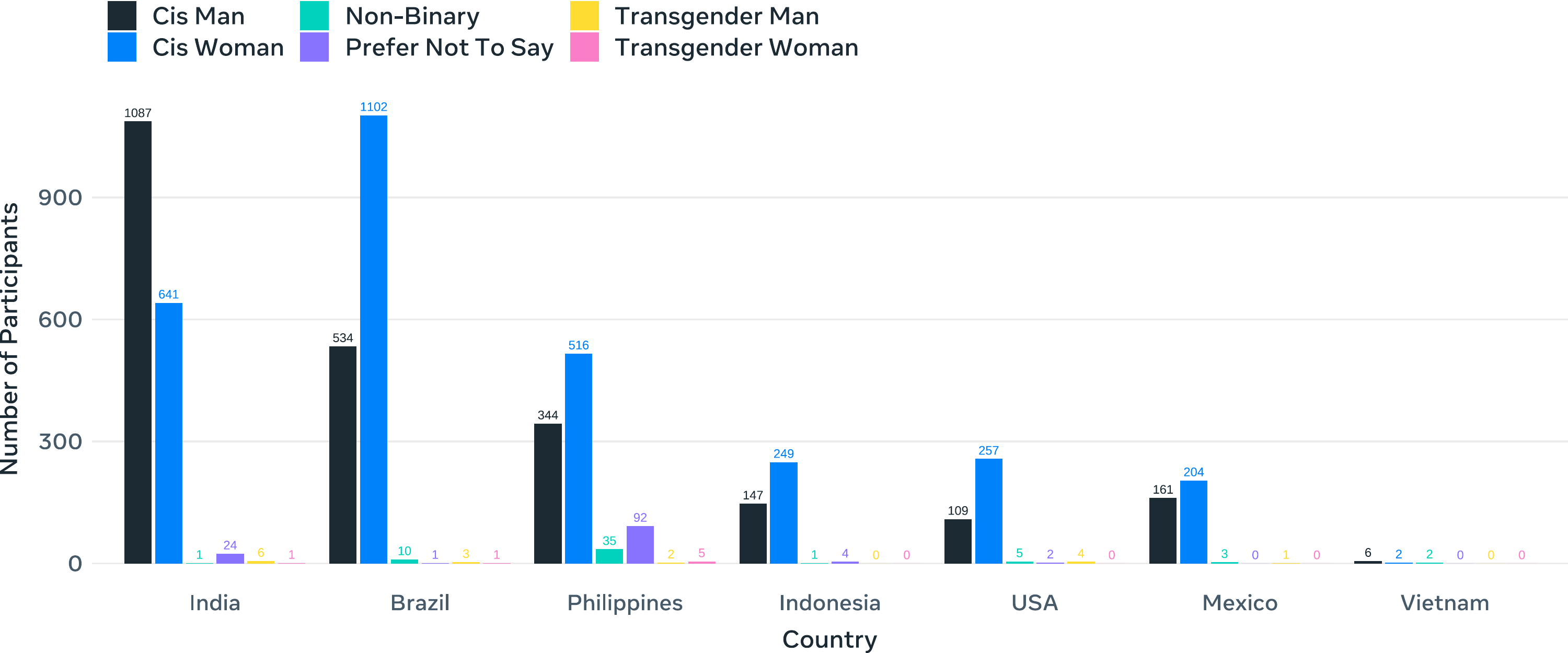}
    \caption{\textbf{Gender} distribution of participants by country. India contains a considerably higher participant ratio of cis men to cis women compared to the other countries.}
    \label{fig:gender}
\end{figure}

In this section, we review a few datasets that were assembled for the purpose of or were later used to measure AI fairness.
The UTKFace dataset~\cite{zhifei2017age} contains over 20,000 face images that were assembled from either other existing datasets or the web, and it contains a wide span of ages (0-116), however, binary gender and five ethnicity groups.
This dataset was primarily assembled to train an age progression/regression model, however, it has been used for evaluating fairness of age, gender, and race detectors~\cite{krishnan2020understanding, lin2022fairgrape}.
FairFace~\cite{karkkainen2021fairface} is a crowd-sourced dataset containing 108,501 images, that are labelled with age, gender and race. The seven race groups used are White, Black, Indian, East Indian, Southeast Asian, Middle East, and Latino, and the dataset is reasonably balanced across these groups. However, race is seen as a social construct~\cite{roth2016multiple} and and its use in categorization exercises may be problematic~\cite{hazirbas2022ccv2lit}.
FairFace also contains age, but non-inclusive binary gender labels.
The authors examined models on four datasets, and reported that models trained on FairFace are more accurate on other datasets, and the accuracy is consistent across race and gender groups.

The MIAP (More Inclusive Annotations for People)~\cite{schumann2021step} is a subset of the OpenImages dataset~\cite{kuznetsova2020open}, and contains additional labels for 100,000 images.
The additional labels provide bounding boxes and demographic attributes for everyone present in an image. 
The demographic attributes include perceived gender and age range. However, the groups for gender (feminine, masculine, unknown) and age range (young, middle, older, unknown) are quite limited in number.
These annotations allow fairness analysis on models performing vision tasks.
RFW~\cite{Wang_2019_ICCV} is a face recognition dataset composed of four race groups (Caucasian, Indian, Asian, African) each of which contains around 10,000 images. In addition to race being a social construct, the number of race groups (4) is also extremely limited.
This dataset was created by pulling images from the MS-Celeb-1M~\cite{ms1_celeb} dataset.
The only annotation provided is race, which is done semi-automatically, where authors check low confidence score of the tool used to predict race.
The authors report that there is a significant bias in verification accuracy towards Caucasians, when compared to the other three race groups. 
And as previously mentioned, our dataset is consent-driven.

MORPH~\cite{ricanek2006morph} is a dataset composed of 55,134 mug-shot photos of 13,617 subjects, where the race, gender (binary), and age attributes are manually assigned for each individual. This dataset also has the recurring issues with race and binary gender.
MORPH was initially assembled to study age progression across several types of facial analysis, including animation, face modeling, and face recognition, however, as it contains manually annotated demographic attributes, this dataset has become commonly used to study fairness of AI models~\cite{drozdowski2020demographic, albiero2020does, albiero2021gendered, qiu2021does, wu2022face}.

\begin{figure}[t!]
    \includegraphics[width=\columnwidth]{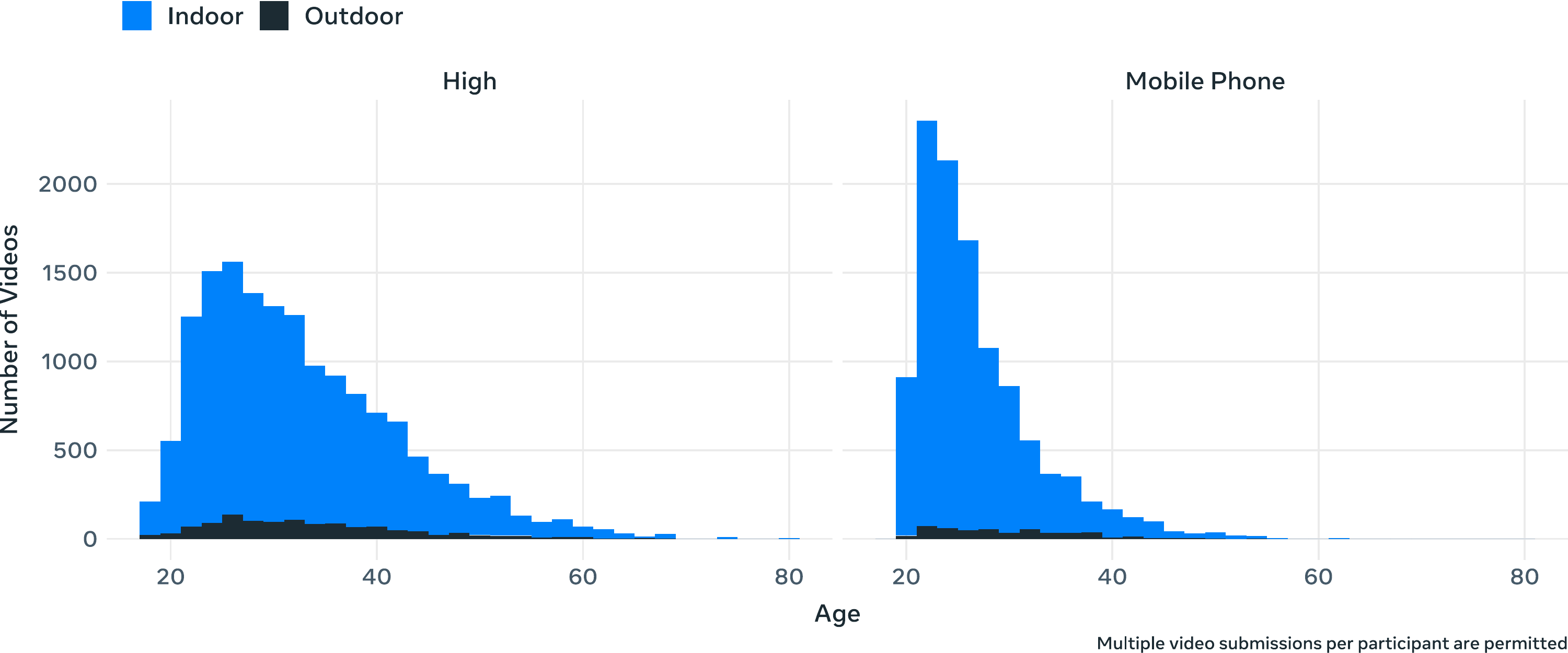}
    \caption{\textbf{Age} distribution by video quality. Most videos are recorded indoors. High quality videos are greater than or equal to 720p.}
    \label{fig:age_quality}
\end{figure}

Differing from the aforementioned datasets, the precursor of the proposed dataset, Casual Conversations (CCv1)~\cite{hazirbas2022ccv1} was created with full consent of the participants.
CCv1 is composed of around 45,000 videos and 3,011 subjects, which were collected across five U.S. cities.
This dataset provides self-provided annotations for age, and gender, and annotated labels for apparent skin tone, and low ambient lighting.
Where previous datasets only use binary genders (with/without unknown as option), which can be seen as discriminative, CCv1 has a multi-choice for gender that are ``male'', ``female'' and ``other''.
This dataset was primarily used in the DeepFake Detection Challenge (DFDC)~\cite{dolhansky2020dfdc}, where videos were augmented with deepfakes, and competitors designed algorithms to detect fake videos.
Alongside fairness for age, gender and apparent skin tone detectors, Casual Conversations~\cite{hazirbas2022ccv1} also report on fairness of DeepFake detectors.

DollarStreet~\cite{rojas2022dollarstreet} dataset is designed to ensure computer vision fairness across different populations, and is composed of images of everyday household items from 63 different countries around the globe. This dataset is composed of 38,479 images with manual annotations for the objects present, containing 289 categories.
Along with these annotations, the dataset also provides country, region, and monthly income as demographic features.
In their experiments with current vision models, the authors show that there is a significant performance bias to household items coming from higher incomes.
They also show that by fine-tuning models on their dataset, fairness issues can be mitigated.

    
\section{Casual Conversations v2 Dataset}
\begin{figure}[t!]
    \includegraphics[width=\columnwidth]{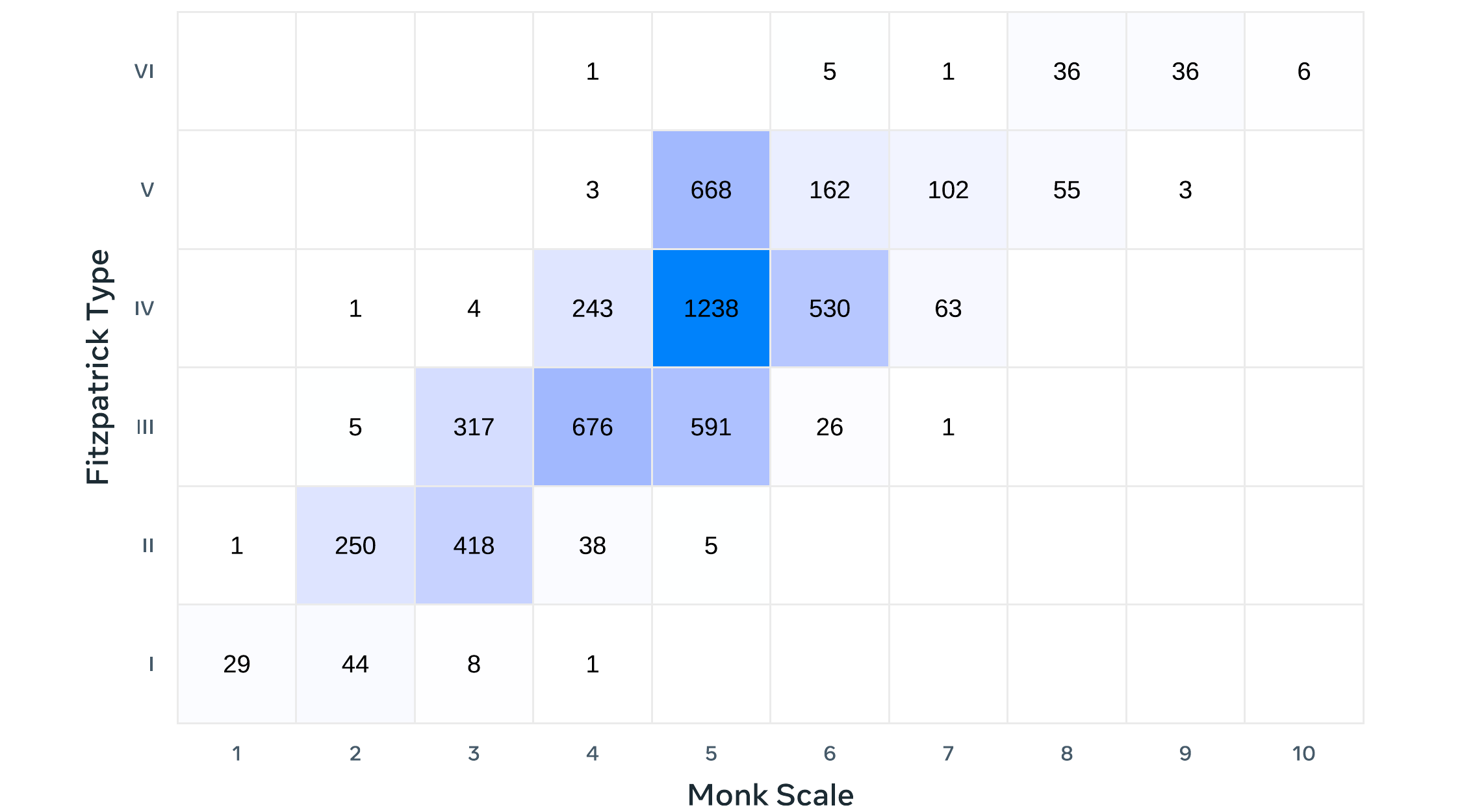}
    \caption{\textbf{Fitzpatrick} vs \textbf{Monk} skin tone scales in CCv2. More than 90\% of the annotations have medium to high confidence in either scale.}
    \label{fig:skintone}
\end{figure}

CCv2 is a dataset that has been carefully curated to improve the fairness and robustness of audio, vision, and speech models. The dataset is diverse across multiple axes, including geographically, demographically, and linguistically, which aids in ensuring that the models that are benchmarked or trained on it are representative of a wide range of subgroups. CCv2 is constructed with the recordings of paid participants that have explicitly provided their consent for their videos' use in research. This ensures that the dataset aligns with ethical standards~\eg ~\cite{jerone2023ethical},  for data collection, respects the privacy and autonomy of the participants, but also promotes transparency and other key ethical considerations in responsible data collection practices.
The dataset offers 7 self-provided and 4 annotated attributes. Categories and subcategories in our dataset were informed by the literature review presented by Hazirbas~\etal~\cite{hazirbas2022ccv2lit}. We collected our dataset in 7 countries,~\ie  Brazil, India, Indonesia, Mexico, Vietnam, Philippines and the United States of America. The dataset is composed of 26,467 videos of 5,567 unique paid participants. There are approximately 674 hours of recordings in total, with 354 hours of videos with \textit{nonscripted} and 320 hours of videos with \textit{scripted} text (see~Appendix~\ref{app:video_setup}).

Participants self-provided their \textit{age, gender, language/dialect, disability, physical adornments \& attributes and geo-location}. For geo-location, we release only ``country'' and ``state/region'' information. All self-provided fields were optional during recordings and participants are allowed to withdraw their data anytime after the collection.

\begin{figure}[t!]
    \includegraphics[width=\columnwidth]{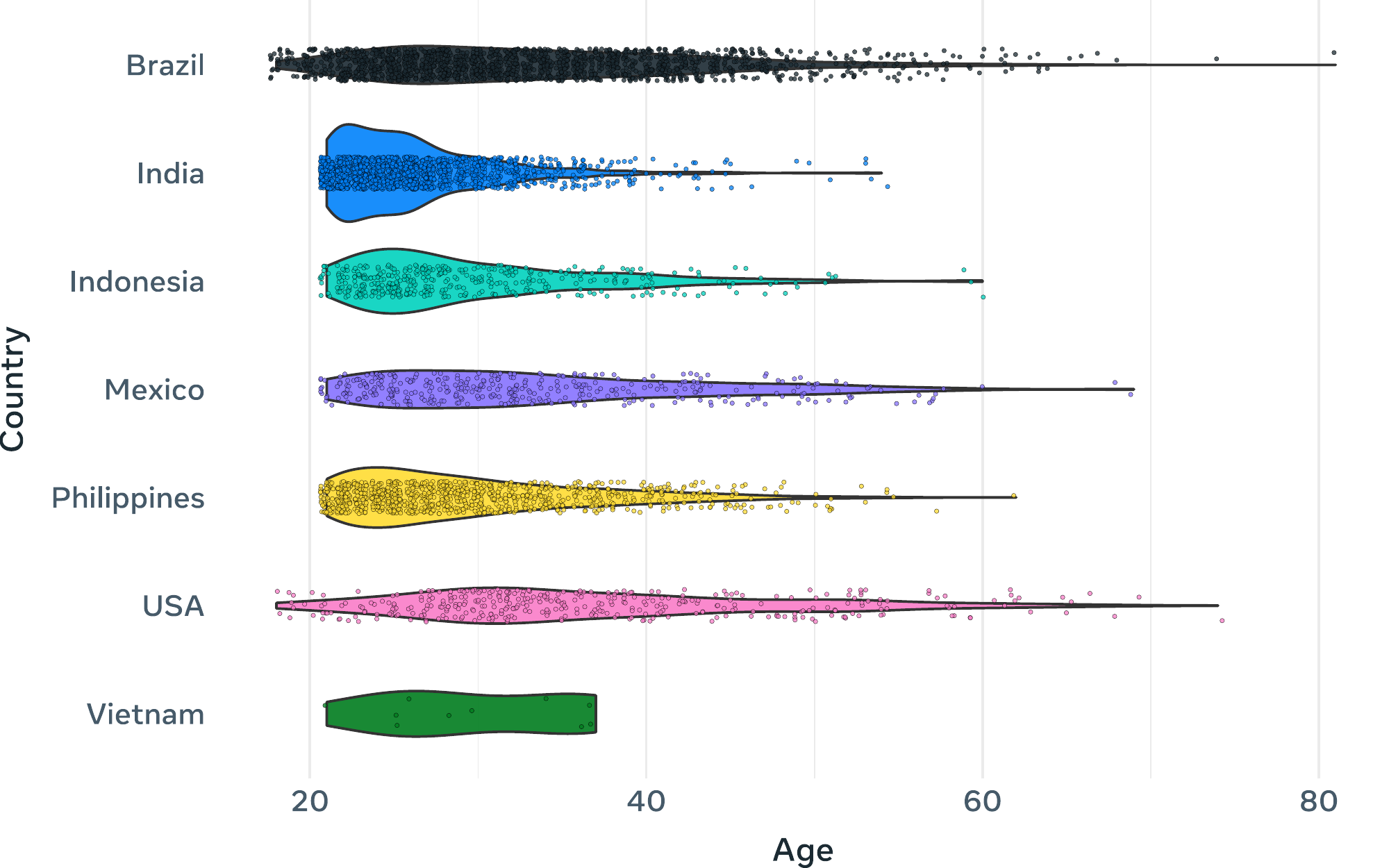}
    \caption{\textbf{Age} distribution by country. The median age of participants in India is 25 compared to 34 in the USA.}
    \label{fig:age_country}
\end{figure}

In addition to the self-provided categories, trained annotators labeled each participant for \textit{Fitzpatrick Skin Type}~\cite{fitzpatrick75soleil} and \textit{Monk Skin Tone}~\cite{monk2014skin} scales, \textit{voice timbre} as well as annotated videos for \textit{different recording setup} and per-second \textit{activity} on videos. By providing both skin tone scales, we believe researchers will be able to also compare these two scales in their studies. \textit{Voice timbre} is usually used in music industry for categorizing singing voice using voice types~\cite{kennedy2023understanding}. In this dataset, we only annotate for \textit{low} (bass), \textit{average} (alto, tenor) and \textit{high} (soprano) pitch. During labelling, we provided an example of voice timbre classification video\footnote{\url{https://www.youtube.com/watch?v=lIfxH2119cU}} and only presented audio to the raters to make the annotation task apparent gender-blind and therefore remove gender-based voice classification bias during labelling.

\begin{figure}[t!]
    \includegraphics[width=\columnwidth]{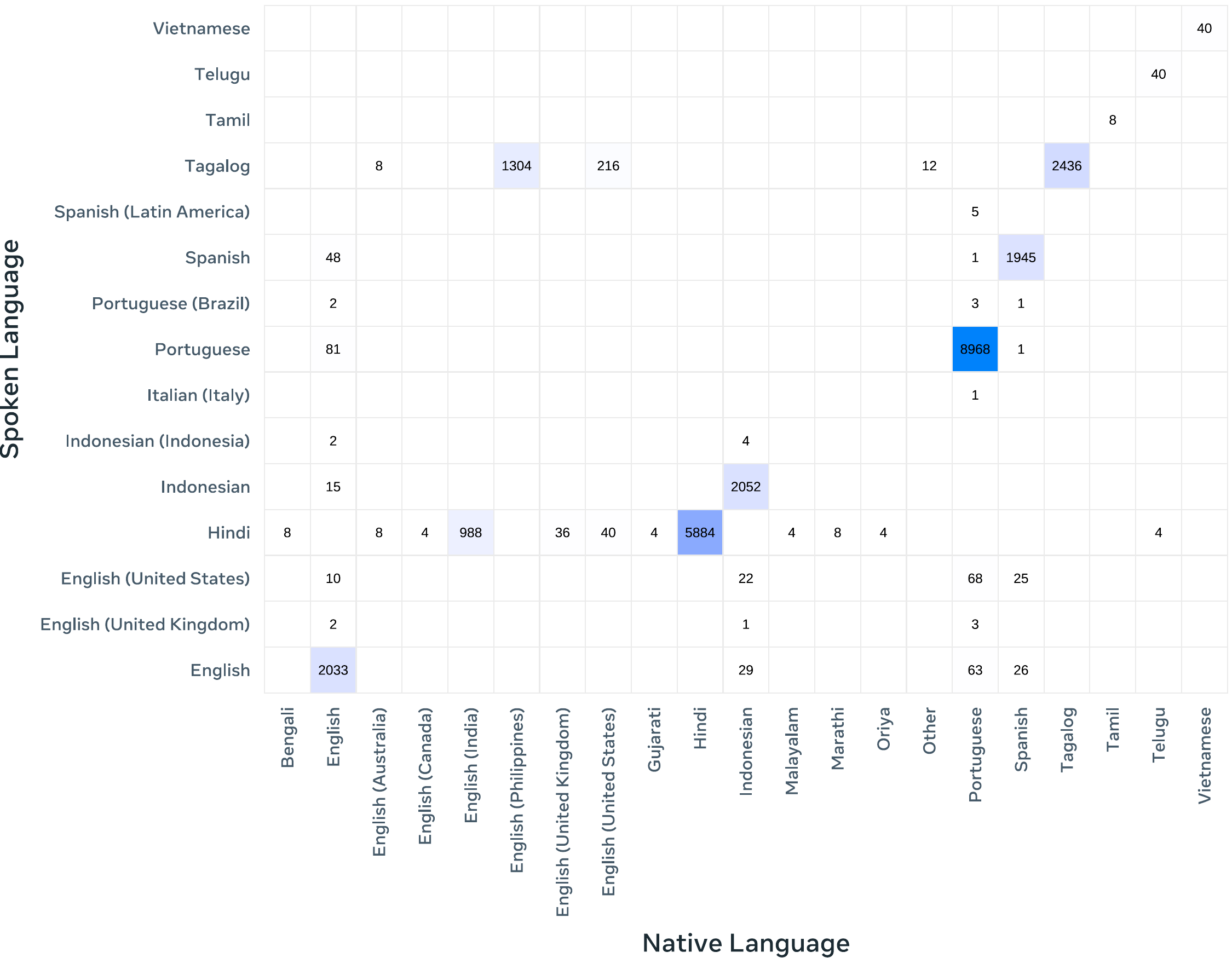}
    \caption{\textbf{Spoken} vs \textbf{Native language/dialect} distribution. Portuguese, Hindi, Indonesian, Tagalog and English are most preferred spoken languages/dialects. Participants may speak in more than one language across their video submissions.}
    \label{fig:language}
\end{figure}

As different recording setups, we include \textit{\textbf{video setup}}: \textit{nonscripted} refers to one of the five questions in Appendix~\ref{app:nonscripted} selected by participants in each \textit{nonscripted} video. Participants may have multiple videos of different nonscripted questions. \textit{Scripted} refers to three paragraphs that are read by participants. Text is sampled from \textit{The Idiot} book by \textit{F. Dostoyevsky} and is translated from English (U.S.) to all ``spoken languages/dialects'' in the videos (see Appendix~\ref{app:subtext});
\noindent \textit{\textbf{video quality}}: \textit{high} if resolution is higher than or equal to 720 pixels or \textit{mobile phone};
\textit{\textbf{background noise}}: \textit{boolean} if there is any background noise in the video;
\noindent \textit{\textbf{capture environment}}: \textit{indoor} or \textit{outdoor};
\noindent \textit{\textbf{hemisphere}}: \textit{north} or \textit{south};
\noindent \textit{\textbf{weather}}\footnote{We include weather also for videos recorded indoor.}: \textit{cloudy}, \textit{dark}, \textit{rainy}, \textit{sunny} and
\noindent \textit{\textbf{video duration in seconds}}.

\textit{Activity} per-second is annotated for \textit{three} categories: 1) \textit{action}: standing, walking, sitting,  laying, waving, 360 rotation; 2) \textit{gesture}: stretching body, raising hand/leg, moving head; and \textit{appearance}: full body visible, upper body visible, lower body visible, only head visible. We asked some of the participants to complete a full rotation (360 degrees) at the end of the video and it is marked under \textit{action}. In some videos, rotation may not be complete and we left it to users' discretion to remove these parts of the videos. Annotations are completed only for the first frame of each second in videos.

In addition to annotated labels, we also share annotators' label confidence for \textit{skin tone}, \textit{voice timbre} and \textit{activity} in the scale of \textit{low}, \textit{medium} and \textit{high}.

Figure~\ref{fig:gender} shows the gender distribution over countries. We have more participants in Brazil and India than the rest of the countries and most of the participants in the dataset identified themselves as ``cis man'' and ``cis woman''.

The age distribution across video quality is shown in Figure~\ref{fig:age_quality}.
The majority of videos are indoors, and with younger participants.
For mobile phone videos, we see a significant peak around the age of twenty. Furthermore, we show the age distribution by country in Figure~\ref{fig:age_country}.
While there is a wide range of ages, most participants are less than 40 years old.

Figure~\ref{fig:language} shows a contingency table between spoken language and native language.
In this figure, we can see the most correlation between the native language and the spoken language, as expected.
The language spoken by most different native language groups is Hindi, while the native language group that speaks most different languages are English and Portuguese.

Figure~\ref{fig:skintone} compares skin tones annotated with Fitzpatrick Type and Monk Scale.
We can see some correlation between both types of skin tone, however, as expected skin tone is more spread in Monk Scale.

\begin{figure}[t!]
    \includegraphics[width=\columnwidth]{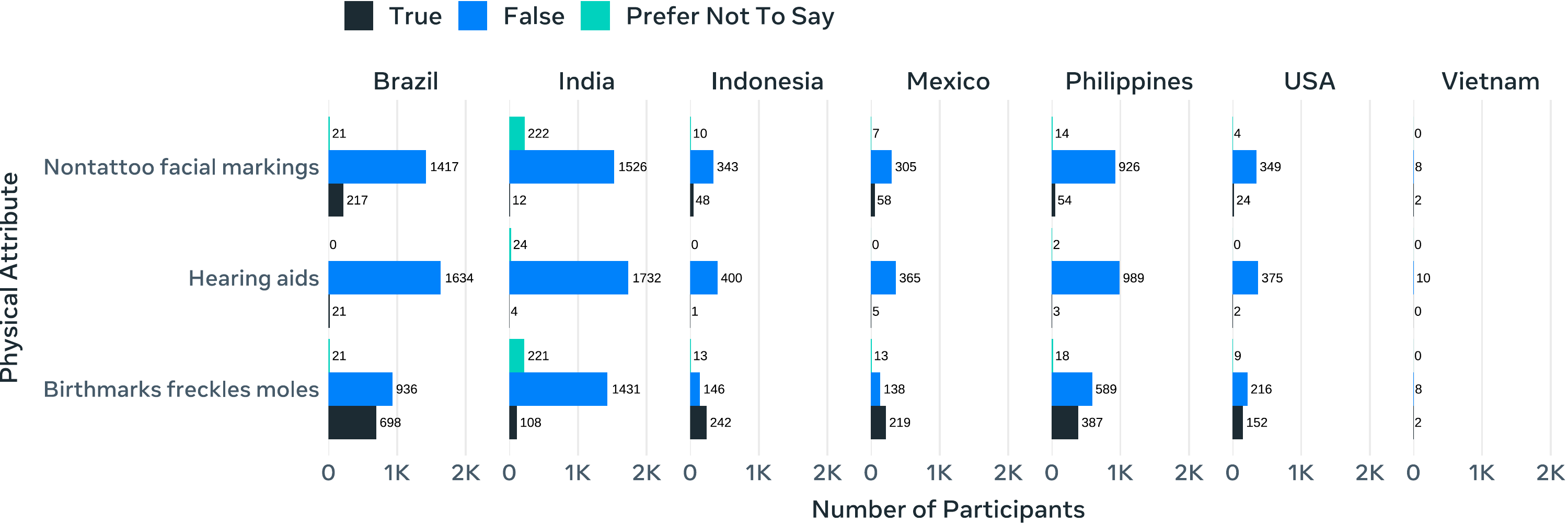}
    \caption{\textbf{Binary physical attributes} distributions by country.}
    \label{fig:binay_phys_attrs}
\end{figure}

\begin{figure}[t!]
    \includegraphics[width=\columnwidth]{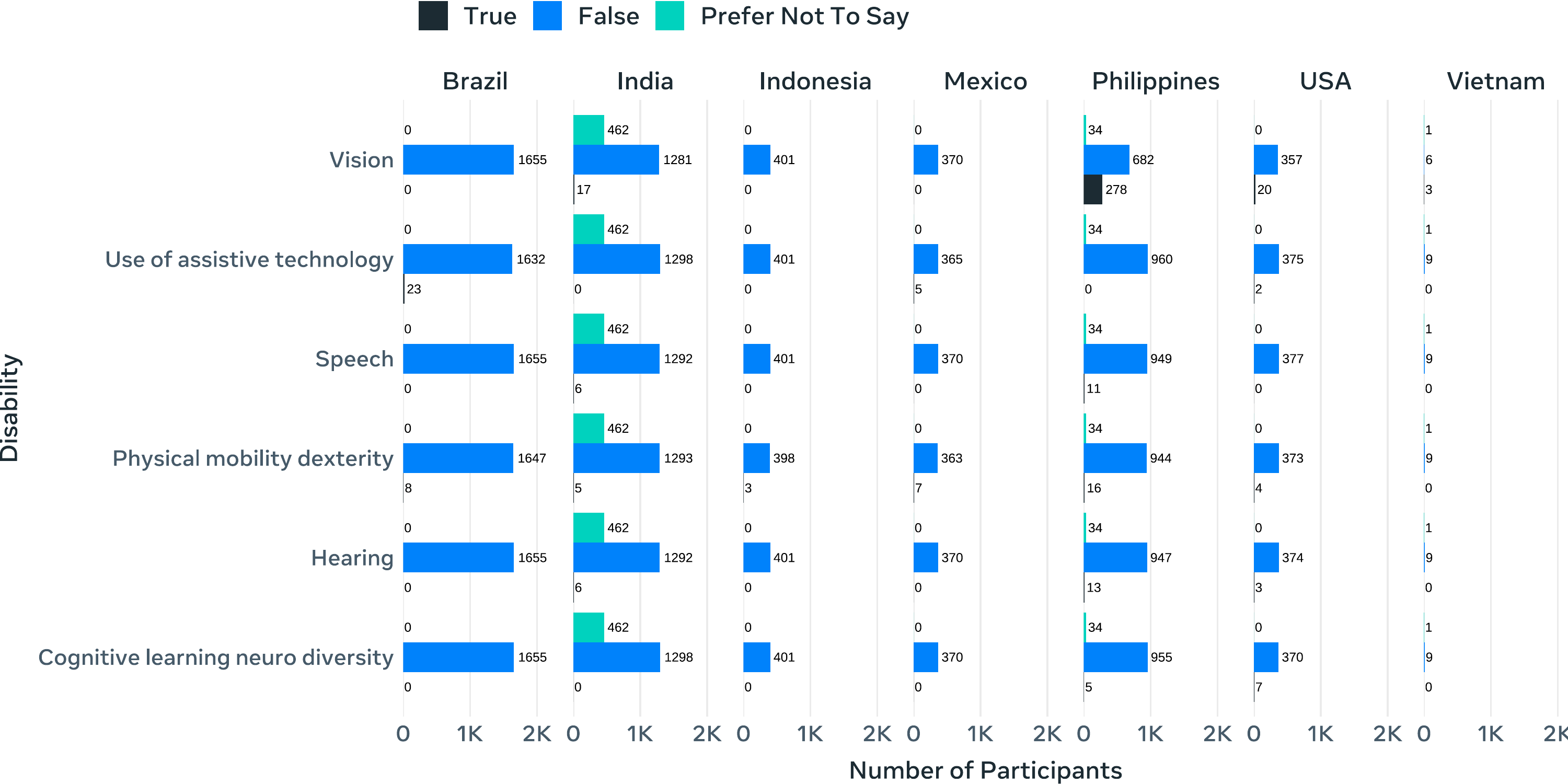}
    \caption{\textbf{Disability} distributions by country.}
    \label{fig:disability_country}
\end{figure}

\begin{figure}[t!]
    \includegraphics[width=\columnwidth]{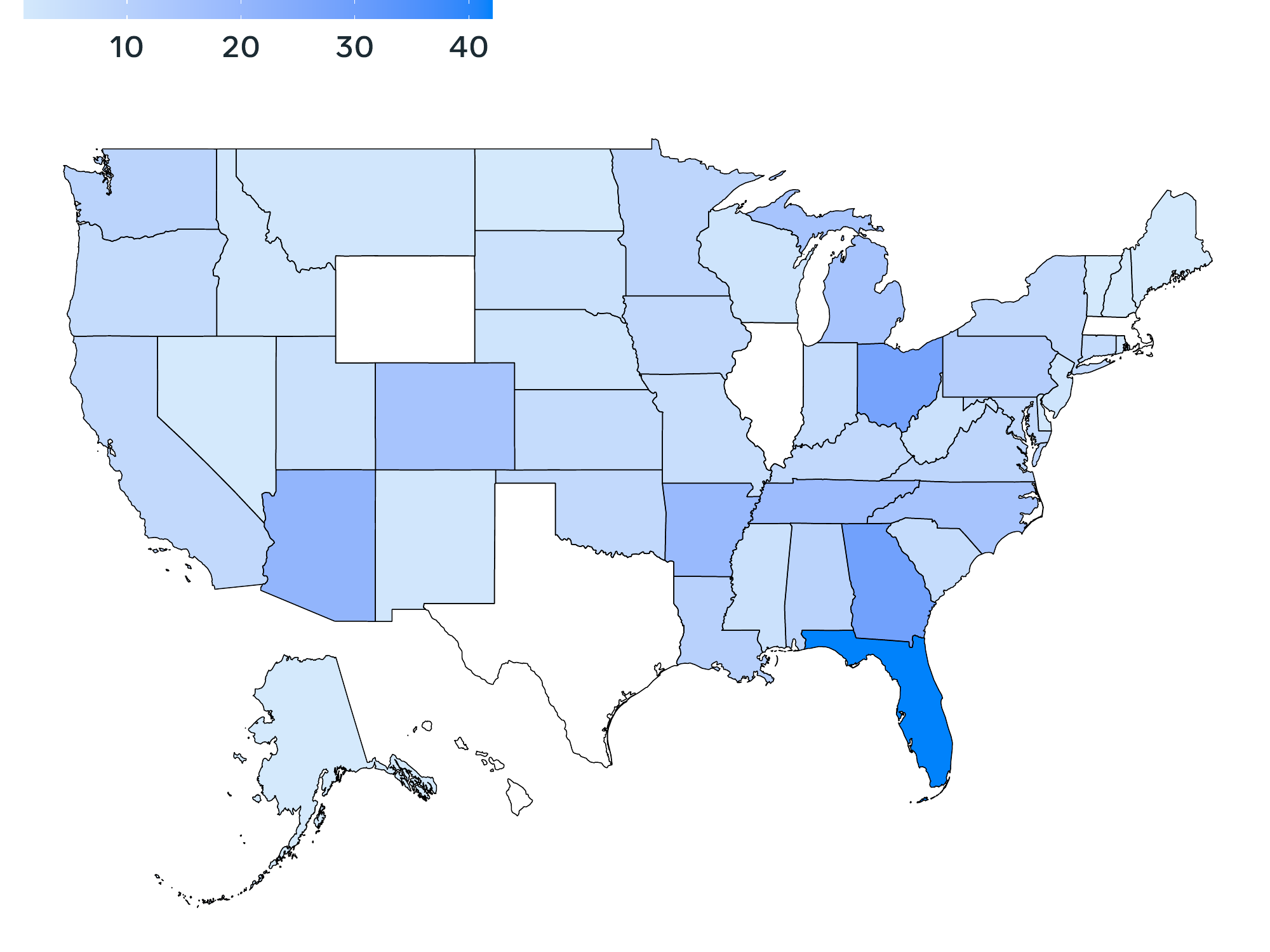}
    \caption{US state distribution of participants.}
    \label{fig:usa_map}
\end{figure}

\begin{figure}[t!]
    \centering
    \includegraphics[width=0.5\columnwidth]{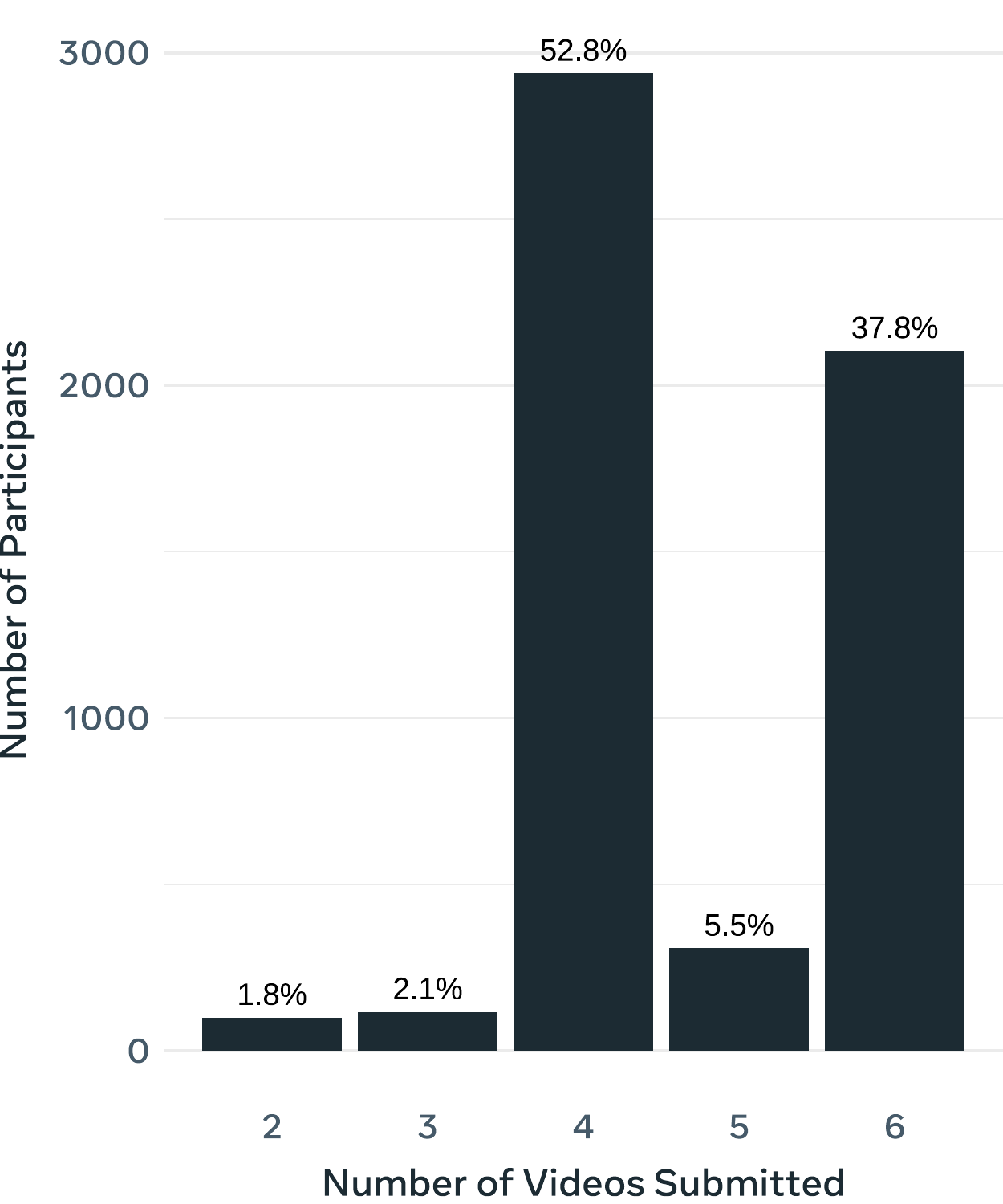}
    \caption{Frequency distribution of uploads per participant.}
    \label{fig:videos_per_subject}
\end{figure}

Furthermore, we also show the binary physical attributes by country in Figure~\ref{fig:binay_phys_attrs}, disability by country in Figure~\ref{fig:disability_country}, US state distribution of participants in Figure~\ref{fig:usa_map} and also the frequency of uploads per participant in Figure~\ref{fig:videos_per_subject}.

\section{Conclusion}
We have presented The Casual Conversations v2 (CCv2) dataset, a diverse, large benchmark for measuring fairness and robustness in audio, vision, and speech models.
To the best of our knowledge, CCv2 is the largest and most diverse consent-driven dataset for fairness and robustness benchmarking.
It contains 7 self-provided and 4 annotated attributes, that enable fairness and robustness measurements across multi-modalities.
It is our hope that CCv2 will boost the development of AI models that are more fair and robust across the proposed attributes.

\noindent \textbf{Acknowledgements}. We would like to extend our heartfelt gratitude to the following teams for their invaluable partnership and support throughout this research: the Civil Rights team, the Accessibility team, the Responsible AI team, the AI Analytics team, the Assistant team, the Speech Recognition team, and the FAIR team. Without their collaboration, this project would not have been a success.
 
We are particularly grateful for the exceptional work and invaluable insights provided by Parisa Assar, who served as the program manager and played a critical role in the project. Her attention to detail, expertise, and dedication were essential to the success and efficient execution of this research.
 
We also extend our appreciation to Ida Cheng for her diligent work. Her commitment to quality has been instrumental in ensuring high standards for the accuracy and reliability of the data we collected.
 
Additionally, we would like to acknowledge the partnership and support provided by Miranda Bogen and Lauren Cohen. Their contributions and insights were instrumental in shaping this dataset.
 
Thank you all for your commitment to this project and for your tireless efforts in advancing our research. Our research has been significantly shaped by your valuable contributions.

{\small
\bibliographystyle{ieee_fullname}
\bibliography{bibliography}
}

\newpage
\appendix
\section{Casual Conversations v2}
In this appendix, we provide details of the categories and subcategories.
\subsection{Annotations File}
We provide annotations in a JSON file. For each video we provide all attributes except activity that is stored in a separate JSON file to allow fast loading of the data into memory. Subject ids are numbered from $0000$ to $5566$ and video names are constructed from \textit{subject id}, \textit{spoken language}, \textit{video setup} and \textit{video number}.

\begin{lstlisting}[language=json,firstnumber=1]
{
    "video_name": "3828_spanish_nonscripted_1.mp4",
    "subject_id": "3828",
    "age": 22,
    "gender": "transgender man",
    "native_language": "spanish",
    "secondary_languages": [
        "english (united states)",
        "english (united kingdom)",
        "spanish (latin america)",
        "japanese (japan)",
        "italian (italy)"
    ],
    "disabilities": {
        "vision": false,
        "hearing": false,
        "physical_mobility_dexterity": false,
        "speech": false,
        "cognitive_learning_neuro_diversity": false,
        "use_of_assistive_technology": false
    },
    "physical_adornments": {
        "have_hair_cover": false,
        "hair_color": "blue",
        "have_beard_mustache": false,
        "have_face_covering": false,
        "have_face_mask": false,
        "have_make_up": false,
        "have_eye_wear": true,
        "have_ear_wear": false,
        "have_visible_tattoos": false,
        "have_bindi": false,
        "have_visible_piercings": false
    },
    "physical_attributes": {
        "hair_type": "curly",
        "hair_color": "brunette",
        "have_hearing_aids": false,
        "eye_color": "brown",
        "have_birthmarks_freckles_moles": true,
        "have_nontattoo_facial_markings": false
    },
    "voice_timbre": {
        "type": "high pitch",
        "confidence": "high"
    },
    "fitzpatrick_skin_tone": {
        "type": "type iii",
        "confidence": "medium"
    },
    "monk_skin_tone": {
        "scale": "scale 5",
        "confidence": "medium"
    },
    "geo_location": {
        "country": "mexico",
        "state_region": "baja california"
    },
    "video_setup": {
        "type": "nonscripted",
        "speech_topic":
        "tell us how you usually spend your weekends"
    },
    "video_quality": "high",
    "background_noise": false,
    "spoken_language": "spanish",
    "capture_environment": "indoor",
    "hemisphere": "northern_hemisphere",
    "weather": "sunny",
    "video_duration_secs": "91.895122"
}
\end{lstlisting}

We recommend users to load the JSON files into memory using Pandas library in Python:

\begin{lstlisting}[language=python,firstnumber=1]
# import Pandas
import pandas as pd

# load annotations into memory
annotations = pd.read_json("CasualConversationV2.json", dtype={'subject_id': object})

# load activity annotations in memory
activity_annotations = pd.read_json("CasualConversationV2_activity.json", dtype={'subject_id': object})

# serialize the dictionary fields
geoloc = annotations.geo_location.apply(pd.Series)

# concatenate country with the data frame
annotations = pd.concat([annotations, geoloc.country], axis=1)
\end{lstlisting}

\subsection{Video Setup}
\label{app:video_setup}
In video setup, for each participant we have at least one nonscripted and one scripted video. Most of the participants recorded more than two videos. Exact total duration of recordings is 674 hrs 22 min 45 sec. Out of this, 354 hrs 28 min 8 sec of it with nonscripted and 319 hrs 54 min 37 sec of it with scripted text.

\subsubsection{Nonscripted Text}
\label{app:nonscripted}

In nonscripted videos, we asked participants to choose one of the following five questions and provide a monologue in about 1 minute. For each nonscripted video, we store the corresponding question.

\begin{itemize}
    \item[$\ast$] tell us how you connect with family and friends
    \item[$\ast$] tell us how you usually spend your weekends
    \item[$\ast$] tell us if you would rather spend your leisure time in nature or in a city and why
    \item[$\ast$] tell us what activities you like to do in summer
    \item[$\ast$] tell us what you think about the weather in the city you live in
\end{itemize}

\subsubsection{Scripted Text}
\label{app:subtext}

\begin{tcolorbox}[width=\linewidth, colback=green!5, colframe=green!40!black, title=\textit{The Idiot} book by \textit{F. Dostoyevsky}]
“Toward the end of November, during a thaw, at 9 o’clock one morning, a train on the Warsaw and Petersburg railway was approaching the latter city at full speed. The morning was so damp and misty that it was only with great difficulty that the day succeeded in breaking; and it was impossible to distinguish anything more than a few yards away from the rail car windows.\\

Some of the passengers by this particular train were returning from abroad; but the third-class carriages were the most filled up, mainly with insignificant persons of various occupations and degrees, picked up at the different stations nearer town. All of them seemed weary, and most of them had sleepy eyes and a shivering expression, while their complexions generally appeared to have taken on the color of the fog outside.”\\

One of them was a young man of about twenty-seven, not tall, with black curling hair, and small, gray, fiery eyes. He wore a large fur—or rather astrakhan—overcoat, which had kept him warm all night, while his neighbor had been obliged to bear the full severity of a Russian November night entirely unprepared. The wearer of this cloak was a young man, also of about twenty-six or twenty-seven years of age, slightly above average height, very fair, with a thin, pointed and very light-colored beard; his eyes were large and blue, and had an intent look about them.\\

\noindent “Cold?”\\

“Very,” said his neighbor, readily, “and this is a thaw, too. Imagine if it had been a hard frost! I never thought it would be so cold in the old country. I’ve gotten quite unaccustomed to it.”\\

“What, been abroad, I suppose?”\\

“Yes, straight from Switzerland.”\\

“Wow! My goodness!” The young, black-haired man whistled, and then laughed.

\end{tcolorbox}

The scripted text is taken from \textit{The Idiot} book by \textit{F. Dostoyevsky}. The text is translated into all other \textit{twelve} spoken languages/dialects present in the videos and translations are released as part of the annotations. All ``spoken languages/dialects'' are English (U.S.), English (U.K.), Hindi, Indonesian, Italian, Portuguese, Portuguese (Brazil), Spanish, Spanish (Latin America), Tagalog, Tamil, Telugu and Vietnamese.

\end{document}